\documentclass{bmvc2k}

\usepackage{multirow}

\title{KAN-Mamba FusionNet: Redefining Medical Image Segmentation with Non-Linear Modeling}

\addauthor{Akansh Agrawal$^{*}$}{akanshcs2020@gmail.com}{1}
\addauthor{Akshan Agrawal$^{*}$}{akshancs2020@gmail.com}{1}
\addauthor{Shashwat Gupta}{guptashashwatme@gmail.com}{1}
\addauthor{Priyanka Bagade$^\dag$}{pbagade@iitk.ac.in}{1}

\addinstitution{
Indian Institute of Technology Kanpur \\
Kanpur, Uttar Pradesh, India
}

\runninghead{A. Agrawal, A. Agrawal, S. Gupta, P. Bagade}{KAN-Mamba FusionNet}

\begin{document}

\maketitle

\begin{abstract}
Medical image segmentation is essential for applications like robotic surgeries, disease diagnosis, and treatment planning. Recently, various deep-learning models have been proposed to enhance medical image segmentation. One promising approach utilizes Kolmogorov-Arnold Networks (KANs), which better capture non-linearity in input data. However, they are unable to effectively capture long-range dependencies, which are required to accurately segment complex medical images and, by that, improve diagnostic accuracy in clinical settings. Neural networks such as Mamba can handle long-range dependencies. However, they have a limited ability to accurately capture non-linearities in the images as compared to KANs. Thus, we propose a novel architecture, the KAN-Mamba FusionNet, which improves segmentation accuracy by effectively capturing the non-linearities from input and handling long-range dependencies with the newly proposed KAMBA block. We evaluated the proposed KAN-Mamba FusionNet on three distinct medical image segmentation datasets: BUSI, Kvasir-Seg, and GlaS - and found it consistently outperforms state-of-the-art methods in IoU and F1 scores. Further, we examined the effects of various components and assessed their contributions to the overall model performance via ablation studies. The findings highlight the effectiveness of this methodology for reliable medical image segmentation, providing a unique approach to address intricate visual data issues in healthcare.
\end{abstract}

\section{Introduction}
Medical image segmentation can help with the accurate localization of anatomical features, leading to the timely detection of abnormalities and patient treatment. 
Techniques like Convolutional Neural Networks (CNNs) are popularly used to mark individual pixels for segmentation tasks to identify tumor locations, organs, and relevant anatomical features \cite{church2019deep,kevric2022deep,zhang2021medical,yamashita2018cnn}. Classical CNN-based approaches like UNet models \cite{ronneberger2015unet} use the encoder and decoder architecture for image segmentation. 
The UNet variations, UNet \cite{ronneberger2015unet}, UNet++ \cite{zhou2018unetplusplus}, and UNet3+ \cite{huang2020unet3plus} use hierarchical techniques to extract features from the input image. Thus, they fail to get global contextual dependencies, which are essential to segment varied-sized anatomical parts from the medical image. 

Unlike CNNs, ViT, i.e., vision transformers \cite{dosovitskiy2021image} based models like Medical Transformer \cite{valanarasu2021medical} and TransUNet \cite{chen2021transunet} use attention mechanism to capture the global contextual dependencies and can accept varied input sizes, making them better suitable for medical image segmentation. However, transformers suffer from large model sizes, high memory, and computational requirements due to long-run feature extractions. 

In order to overcome the challenges of the higher computational complexity of transformers, state space models (SSM) using linear RNNs (Recurrent Neural Networks) are proposed in Mamba architecture and its variations \cite{gu2023mamba,ma2024umamba,wang2024segmamba,wang2024visionmamba}. These models also address the limitations of CNNs in capturing long-range dependencies. However, these models have conventional convolution layers, which use static activation functions and thereby, limiting their ability to handle non-linearities. 

Recent architectures like U-KAN  \cite{li2024ukan}, employs KAN layers to better capture the non-linear intricacies in the images. The KAN layers use learnable activation functions, to efficiently approximate multivariate continuous functions using univariate transformations, enabling the capture of non-linearities in the input data \cite{liu2024kan}. 
However these architectures cannot capture long-range dependencies. Further, neither U-KAN nor Mamba considers specific attention to local features, which are required to extract fine-grained features to identify the boundaries of irregular shapes and sizes of the anatomical structures. 


In this paper, we propose a novel architecture, KAN-Mamba FusionNet, to accurately segment medical images by handling long-range dependencies and efficiently capturing non-linearities. As part of this architecture, we designed a novel KAMBA block by utilizing State Space Model (SSM), KAN layer, spatial attention module, bag of activation (BoA) functions and a skip connection. Our proposed architecture outperforms the state-of-the-art methods on distinct medical image datasets. The contributions are summarised as follows:
\begin{itemize} 
\item We propose a novel KAMBA block to handle long-range dependencies, while efficiently capturing non-linearities and attention to relevant local features. 
\item We also propose bag of activation (BoA) functions to better capture non-linearities in the KAMBA framework against the use of a pre-defined single activation function in various connections. 
\item We present a KAN-Mamba FusionNet architecture by incorporating KAMBA block in the U-KAN pipeline to improve performance of medical image segmentation.
\item We evaluated our proposed architecture on medical image datasets, BUSI \cite{al2020busi}, GlaS \cite{valanarasu2020kiu} and Kvasir-Seg \cite{jha2020kvasir} and provided ablation study in the end.
\end{itemize}

\begin{figure*}[ht]
    \centering
    \includegraphics[width=1\textwidth]{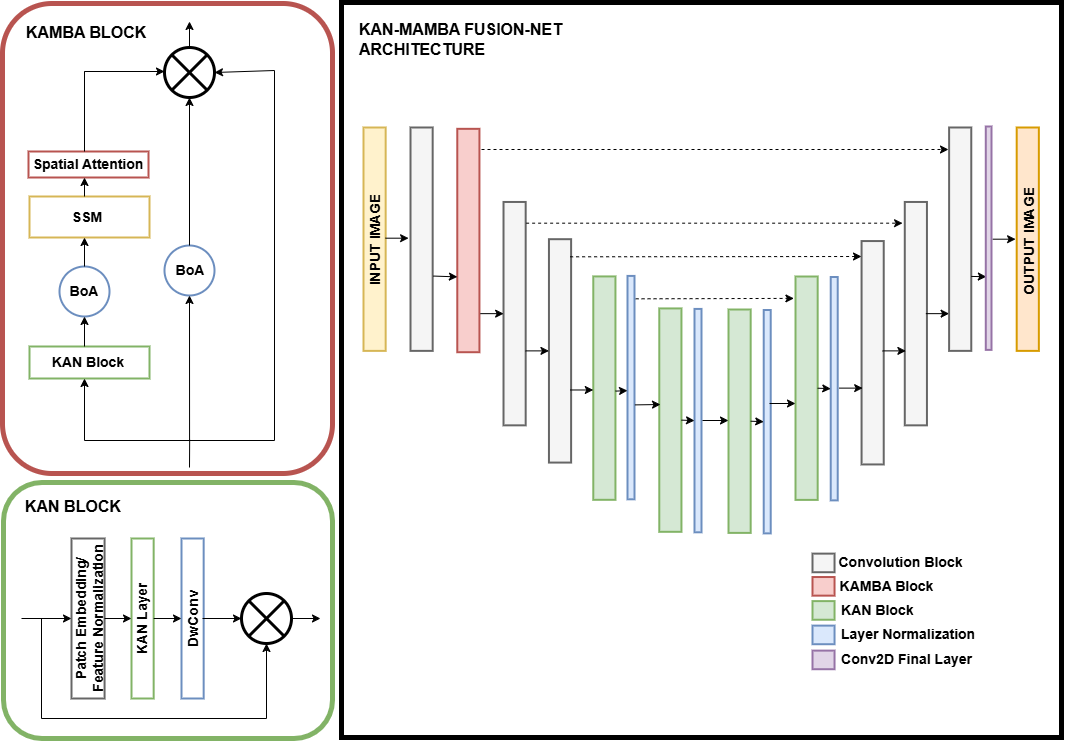} 
    \caption{An overview of the proposed KAN-Mamba FusionNet Architecture with the detailed framework of KAMBA block.}
    \label{fig:overall-image}
\end{figure*}

\section{Methods}
\label{sec:formatting}

Fig. \ref{fig:overall-image} illustrates the proposed KAN-Mamba FusionNet architecture. It consists of three main components: 1) KAN Block, 2) KAMBA Block, and 3) Convolution Block. The proposed architecture incorporates KAMBA block in the U-KAN \cite{liu2024kan} pipeline, as shown in Fig. \ref{fig:overall-image}. The loss for overall architecture is calculated using a combination of binary cross entropy (BCE) loss function and dice loss function as explained in detail in section \ref{sec:implementation details}.

The implementation of the architecture blocks is explained in detail as follows:  

\subsection{KAN Block}
\label{KAN}

Kolmogorov-Arnold Networks (KANs) offer an innovative alternative to traditional MLPs for medical segmentation \cite{li2024ukan,liu2024kan}. KANs efficiently approximate multivariate continuous functions using univariate transformations, enabling the capture of non-linearities in high-dimensional data. KAN layers have demonstrated to better capture non-linearities from the input data by using learnable activation functions on the weight edges connecting two layers of the neural network. This makes the neural network more adaptive to the various intricacies of the data, helping in capturing more complex and non-linear patterns. The KAN having $K$ layers is formulated as:
\begin{equation} \text{KAN}(\textbf{Z}) = (\Phi_{K-1} \circ \Phi_{K-2} \circ \cdots \circ \Phi_1 \circ \Phi_0) \textbf{Z}
\end{equation}
  where, $\Phi$ are expressed using learnable activation functions $\phi$, such that,
\begin{equation}
 \Phi = \{\phi_{q,p}\}, \quad p = 1, 2, \dots, n_{\text{in}}, \quad q = 1, 2, \dots, n_{\text{out}}
\end{equation}


 KANs demonstrate promising accuracy and interpretability over other neural networks \cite{liu2024kan}. Taking advantage of these properties offered by KAN layers, we employ KAN blocks in our architecture. As shown in KAN block in Fig. \ref{fig:overall-image}, the input features are passed through a unit of the KAN layer followed by a depthwise convolution $DwConv$ layer.

\subsection{KAMBA Block} As illustrated in Fig. \ref{fig:overall-image}, KAMBA block consists of a KAN block, BoA, State Space Model (SSM), a spatial attention module and a skip connection. These modules help KAMBA block in ensuring long-range dependencies, extracting fine-grained features, while better handling non-linear patterns in the input images. Following subsections explain each module in detail to address how these properties are achieved. \\ \\
\noindent \textbf{KAN Block} As explained in section \ref{KAN}, the KAN block ensures capturing non-linearities in the input data through learnable activation functions. \\  \\
\noindent \textbf{Bag of Activation Functions} 
Different activation functions capture distinct aspects of data, each bringing their own strengths and weaknesses to a neural network. For example, ReLU introduces sparsity by activating only for positive inputs, which can lead to faster convergence and reduced computational complexity. However, ReLU may face the ``dying ReLU'' problem, where neurons may stop learning or become inactive for consistent zero output. On the other hand, the sigmoid function is adept at modeling probabilities and is useful in scenarios requiring output values between 0 and 1. Nonetheless, it is prone to vanishing gradients, which may hamper the training of deep networks. Similarly, the hyperbolic tangent (tanh) activation function captures symmetric relationships around zero, providing outputs in the range of -1 to 1, which may be useful in centering the data and improving convergence. Yet, it also suffers from vanishing gradient issues \cite{lecun1998efficient}.

To overcome these limitations, we propose bag of activation functions (BoA) in our KAMBA block, to replace a single pre-defined activation function in various connections, with a limited set of parameters. BoA is a learnable weighted composition of pre-defined activation functions. 
Mathematically, BoA can be expressed as:
\begin{equation}
\Psi(\textbf{Z}) = \sum_{p=1}^P \alpha_{p} \, \psi_{p}(\textbf{Z}),
\quad where
\sum_{p=1}^P \alpha_{p} = 1
\label{eq:weight_constraint}
\end{equation}

where $P$ is the number of activation functions used, $\alpha_{p}$ forms the learnable weight associated with activation function $\psi_{p}$. For example, the combination can dynamically balance the sparsity introduced by ReLU with the probabilistic modeling of sigmoid and the zero-centered outputs of tanh. This adaptive weighting is achieved through gradient-based optimization, ensuring that the network selects the most appropriate activation behavior for different tasks or data characteristics.

In our BoA layer, we have included ReLU, tanh, Softplus, GELU and SiLU activation functions while keeping the same initial weights. This formulation allows the network to explore a richer function space, enhancing its ability to model diverse patterns in the data. By learning and optimizing the weights associated with each activation function during training, the model can leverage the strengths of each activation function while mitigating their individual weaknesses. \\


\noindent \textbf{State Space Model (SSM)} 
Recent researchers have emphasized the benefits of State Space Models (SSMs) \cite{amo2024,gu2023mamba,ma2024umamba} in addressing the challenge of capturing long-range dependencies, which is crucial for long-sequence based tasks. These models can be viewed as built on a combination of recurrent neural networks (RNNs) and convolutional neural networks (CNNs). These studies have demonstrated that SSMs model the evolution of hidden states through a set of linear recurrence relations for capturing long-range dependencies in input sequences. Moreover, these models are highly efficient at modeling dependencies for the larger segments of the input sequences due to their linear computational complexity in both time and space \cite{gu2023mamba}. Thus, modeling these long-range dependencies can effectively capture global spatial context that can help to accurately segment the relevant features in the input data. As a result, integrating SSM into the KAMBA block helps in efficiently improving the model's segmentation accuracy by considering the long-term relationships within the data. \\

\noindent \textbf{Spatial Attention Module}
The spatial attention module uses inter-spatial relationships among pixels to extract the fine-grained features to efficiently capture the boundaries of the irregular shapes from the input image. Mathematically, spatial attention over an intermediate feature map \textbf{F} can be expressed as \cite{woo2018cbam}:
$$
M_s(\textbf{F}) = \sigma\left(f_{7 \times 7}\left(\left[\text{AvgPool}(\textbf{F}); \text{MaxPool}(\textbf{F})\right]\right)\right)$$ \begin{equation}= \sigma\left(f_{7 \times 7}\left(\left[\textbf{F}_{s\text{avg}}; \textbf{F}_{s\text{max}}\right]\right)\right)
\label{eq:spatial}
\end{equation}

here, $\sigma$ represents the sigmoid function acting over a larger convolution function $f_{7 \times 7}$ with filter size of 7 × 7 to capture more spatial features. \\

\noindent \textbf{Skip Connection} It has been added to avoid the vanishing gradient issues and improve feature extraction by skipping one or more layers from the neural network.  \\

\noindent \textbf{Working of KAMBA Block} The KAMBA block is added after the first convolution block in the KAN-Mamba FusionNet architecture, as shown in Fig. \ref{fig:overall-image}. We positioned the KAMBA block after the first convolution layer to effectively capture the low-level features. This ensures the finer details of the input image are preserved early in the process. 


In KAMBA block, the input feature map ($Z_{in}$) passes through a KAN block ($KANB$). This is further processed by a layer of BoA ($\Psi$), as expressed in equation \eqref{eq:weight_constraint}, followed by an SSM layer and a spatial attention module such that:
\begin{equation}
   \quad  \textbf{Z}_{\text{out}} = KANB(\textbf{Z}_{\text{in}}),
   \quad 
  Z_{\text{out}}' = \Psi(\textbf{Z}_{\text{out}}), \quad
  \textbf{Z}_{\text{out}}'' = M_s(SSM(\textbf{Z}_{\text{out}}')), \quad
\end{equation}

where $M_s(.)$ represents the attention function as explained in equation \eqref{eq:spatial}.

The result of the above formulation is then combined with the output obtained from the parallel BoA layer as well as the initial input fed into the KAMBA block using a skip connection. Mathematically, the final output obtained through KAMBA block can be expressed as:
\begin{equation}
  KAMBA(\textbf{Z}_\text{in}) = \textbf{Z}_{\text{out}}'' \oplus \Psi(\textbf{Z}_{\text{in}}) \oplus \textbf{Z}_{in}
\end{equation}

Thus, our proposed design of the KAMBA block facilitates the integration of SSM, spatial attention and KAN layers to improve the contextual understanding of the relevant anatomical structures in the input data. 


\subsection{Convolution Block} The Convolution block consists of 2D convolution layers, batch normalization layers, a max pooling layer and ReLU activation functions. These modalities help in feature extraction from the input data.




\section{Datasets}
We evaluated our model on three distinct publically available medical image segmentation datasets: Breast UltraSound Images (BUSI) \cite{al2020busi}, Segmented Polyp Images (Kvasir-Seg) \cite{jha2020kvasir} and Gland Segmentation Images (GlaS) \cite{valanarasu2020kiu}. Given the unique characteristics of each dataset, these evaluations offer a strong support for testing the effectiveness of our method. These datasets contain input images as well as the corresponding ground truth masks, which are manually annotated by the medical experts and serve as reference standards for evaluating model performance.

\noindent \textbf{BUSI} \cite{al2020busi}: The dataset consists of ultrasound scans, along with their corresponding segmentation masks for identifying breast cancer-related tumors. The entire set consists of 708 images, out of which 210, 437 and 133 represent the respective number of images for malignant, benign and normal breast cancer cases. We
utilized the images representing breast cancer. The images were uniformly resized to 256 × 256 pixels. 

\noindent \textbf{Kvasir-Seg} \cite{jha2020kvasir}: The dataset consists of 1000 gastrointestinal polyp images (polyps are precursors to colorectal cancer) and their corresponding segmentation masks. 
All images were uniformly resized to 256 × 256 pixels. 

\noindent \textbf{GlaS}  \cite{valanarasu2020kiu}: The dataset consists of gland segmentation images and is associated with the Hospital Clinic in Barcelona, Spain. For our study, we used 165 images from the dataset, all resized to 256 × 256 pixels.

\section{Experiments and Results}
\subsection{Implementation Details}
\label{sec:implementation details}
For all three datasets, we set the learning rate to 1e-4 and trained with the Adam optimizer, incorporating a cosine annealing learning rate scheduler with a minimum learning rate of 1e-5. The loss function used was a combination of binary cross-entropy and dice loss. Each dataset was split in a 4:1 ratio for training and validation, respectively. Training was conducted over 400 epochs, with basic data augmentations like random rotation and flipping, applied to the inputs. The model was implemented using PyTorch on an NVIDIA RTX A6000 GPU. 

The loss in KAN-Mamba FusionNet architecture is computed using a combination of binary cross entropy (BCE) loss and dice loss functions, measuring the deviation from ground truth labels for entire set of N pixels. These losses are defined in \eqref{bce} and \eqref{dice} as follows:

\begin{equation}
\text{Loss}_{\text{BCE}}(\hat{z},z) = -\frac{1}{N} \sum_{i=1}^{N} \left( z_i \cdot \log(\hat{z}_i) + (1 - z_i) \cdot \log(1 - \hat{z}_i) \right)
\label{bce}
\end{equation}

\begin{equation}
\text{Loss}_{\text{Dice}}(\hat{z},z) = 1- \frac{2 \sum_{i=1}^{N} \hat{z}_i \cdot z_i + \text{c}}{\sum_{i=1}^{N} \hat{z}_i + \sum_{i=1}^{N} z_i + \text{c}}
\label{dice}
\end{equation}

where $z_i$ and $\hat{z}_i$ represent the pixel-wise label from actual
and estimated measures respectively. A small constant $c$ is included in dice loss to handle division by zero. The BCE loss quantifies the pixel-wise classification error, whereas the dice loss is useful for obtaining accurate segmentation results, since it emphasizes region-based optimization. The dice loss also handles class imbalance in datasets by measuring the overlap between estimated and ground-truth masks. The combination of these two losses, as defined in equation \eqref{eq:loss}, helps in faster and more consistent convergence, leading to more accurate clinically relevant segmentation results. 

\begin{equation}
\text{Loss}_{\text{Final}} = 0.5*\text{Loss}_{\text{BCE}} + \text{Loss}_{\text{Dice}}
\label{eq:loss}
\end{equation}

\subsection{Evaluation Details} 
\label{sec:evaluation details}
To evaluate the model's performance, we reported validation IoU (Intersection over Union) and F1 Score on all the datasets and compared with the state-of-art methods. The IoU measures the overlap between predicted and ground truth segments, while the F1 score provides a balanced measure of precision and recall. These metrics are crucial for understanding the model’s accuracy and reliability across various segmentation tasks. To ensure fairness and reliability, all performance comparisons with SOTA models were conducted on the same GPU, NVIDIA RTX A6000, with three independent runs. The reported results are averages from the three independent runs. We also present computational cost metrics such as GFLOPs and the total number of model parameters, and compare these with state-of-the-art methods. Additionally, we performed ablation studies of our model to study the effect of our proposed methods. 

\subsection{Performance Comparisons With State-of-Art Methods}

\label{sec:sota}
\begin{table*}[t]
\centering
\fontsize{7.8}{12}\selectfont
\begin{tabular}{|c|c|c|c|c|c|c|}
\hline
\multirow{2}{*}{Method} & \multicolumn{2}{c|}{BUSI \cite{al2020busi}}   & \multicolumn{2}{c|}{Kvasir-Seg \cite{jha2020kvasir}} & \multicolumn{2}{c|}{GlaS \cite{valanarasu2020kiu}} \\ \cline{2-7} 
                                & IoU↑&F1↑&IoU↑&F1↑&IoU↑&F1↑ \\ \hline
U-Net \cite{ronneberger2015unet} &57.63$\pm$0.43 & 72.21$\pm$0.11&72.41$\pm$0.37&83.43$\pm$1.12&82.58$\pm$0.63&90.47$\pm$0.45\\
U-NeXt \cite{valanarasu2022unext} &59.71$\pm$0.31& 73.41$\pm$0.34 & 69.36$\pm$0.26 & 81.50$\pm$0.14 & 82.74$\pm$0.45 & 90.72$\pm$0.32\\
Rolling-UNet \cite{liu2024rolling}& 61.57$\pm$0.27& 74.51$\pm$0.04 & 75.51$\pm$1.07 & 85.71$\pm$0.42 & 84.21$\pm$0.74 & 91.43$\pm$0.78\\
U-Mamba \cite{ma2024umamba} & 61.75$\pm$0.52& 74.91$\pm$0.18 & 73.91$\pm$0.61 & 85.23$\pm$0.52 & 85.12$\pm$0.51 & 91.57$\pm$0.38\\
Seg. U-KAN \cite{li2024ukan} & 63.64$\pm$0.72& 76.68$\pm$0.79 & 75.85$\pm$0.53 & 85.96$\pm$0.35 & 85.45$\pm$0.34 & 91.95$\pm$0.31\\
\rowcolor[rgb]{0.678,0.847,0.902}
\parbox{2cm}{\centering KAN-Mamba FusionNet}& 65.84$\pm$0.54& 79.17$\pm$0.45 & 76.95$\pm$0.68 & 86.66$\pm$0.43 & 85.73$\pm$0.28 & 92.29$\pm$0.23\\
\hline
\end{tabular}
\vspace{2pt} 
\caption{Performance comparison of various segmentation methods on BUSI, Kvasir-Seg and GlaS datasets.}
\label{tab:comparisons}
\end{table*}

Table \ref{tab:comparisons} shows performance comparisons with other state-of-art methods on image segmentation on the three datasets. We introduce comparisons to the five state-of-art methods, U-Net \cite{ronneberger2015unet}, U-NeXt \cite{valanarasu2022unext}, Rolling-UNet \cite{liu2024rolling}, U-Mamba \cite{ma2024umamba}, and Seg. U-KAN \cite{li2024ukan}. Our experimental results demonstrate that, across all datasets, our proposed method, KAN-Mamba FusionNet, consistently performs well in comparison to other state-of-the-art approaches. 

Our model performance, as demonstrated in terms of IoU and F1 score in Table \ref{tab:comparisons}, clearly indicates its robustness and versatility in segmenting various datasets. A consistent high F1 score across datasets indicates the effectiveness of the proposed KAN-Mamba FusionNet to identify true positives while minimizing the number of false positives and false negatives to avoid misdiagnosis.

\begin{figure*}[ht]
    \centering
    \includegraphics[width=1\textwidth]{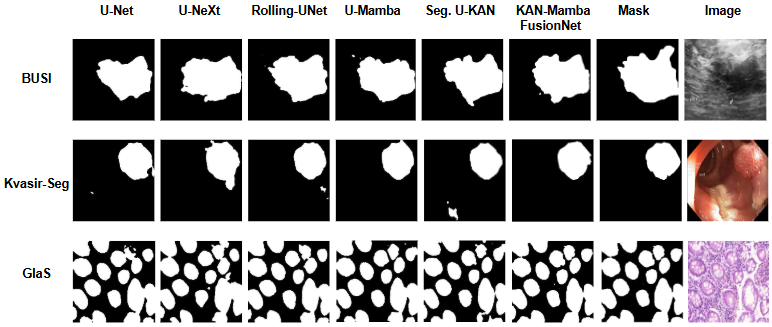} 
    \caption{Visualized segmentation results on three distinct datasets: BUSI, Kvasir-Seg, and GlaS. For each dataset, an input image, the corresponding ground truth mask, and the outputs from SOTA methods and the proposed KAN-Mamba FusionNet model are shown.}
    \label{fig:visualization-image}
\end{figure*}

Fig. \ref{fig:visualization-image} shows the visualization of segmentation results of the proposed KAN-Mamba FusionNet model and SOTA methods on BUSI, Kvasir-Seg and GlaS datasets. The visual outputs highlight that the proposed KAN-Mamba FusionNet model can effectively segment relevant anatomical features for different datasets. The proposed model performs better compared to the SOTA methods in accurately capturing the relevant features, including the boundaries of irregular shapes and sizes in the input images. Also, as can be inferred from these visualized segmentation results, the proposed model exhibits significantly fewer false positive regions compared to the SOTA methods. The closeness of the segmented results produced by the proposed method to the ground truth masks on different datasets strongly suggests the versatility and effectiveness of our model. It is also validated with the consistent high F1 score and IoU as shown in Table \ref{tab:comparisons}.

Additionally, as shown in Table \ref{tab:complexity}, we have calculated the GFLOPs and model parameters to emphasize that our model can be implemented with minimal trade-off in computational cost and performance.

\begin{table*}[t]  
\centering  
\fontsize{8}{12}\selectfont
\begin{tabular}{|c|c|c|}
\hline
\multirow{2}{*}{Method} & \multicolumn{2}{c|}{ Complexity} \\
\cline{2-3}
                       & GFLOPS & Params (M) \\
\hline
U-Net \cite{ronneberger2015unet} & 524.2 & 34.53 \\
U-NeXt \cite{valanarasu2022unext} & 3.98 & 1.47 \\
Rolling-UNet \cite{liu2024rolling} & 13.41 & 1.78 \\
U-Mamba \cite{ma2024umamba}& 2087 & 86.3 \\
Seg. U-KAN \cite{li2024ukan}  & 13.87 & 6.36 \\
\rowcolor[rgb]{0.678,0.847,0.902}
KAN-Mamba FusionNet  & 13.96 & 6.37 \\
\hline
\end{tabular}
\vspace{10pt} 
\caption{Model complexity comparison of various methods.}
\label{tab:complexity} 
\end{table*}

\subsection{Ablation Studies}
\label{sec:ablation}
\begin{table*}[t]
\centering
\fontsize{8}{12}\selectfont
\begin{tabular}{|c|c|c|c|c|c|c|}
\hline
\multirow{2}{*}{Blocks and/or Layers} & \multicolumn{6}{c|}{Performance Metrics} \\
\cline{2-7}
                      & IoU$\uparrow$ & F1$\uparrow$ & Accuracy$\uparrow$ & AUC$\uparrow$ & Precision$\uparrow$ & Recall$\uparrow$ \\
\hline
Classical Mamba            & 60.87          & 75.12         & 92.63              & 91.56         & 73.93             & 72.74\\
Classical Mamba + KAN              & 62.73          & 77.43         & 93.43               & 92.04        & 74.14               & 75.22 \\
Classical Mamba with BoA + KAN              & 64.45          & 77.32         & 93.12             & 93.48         & 76.65               & 74.48 \\
\rowcolor[rgb]{0.678,0.847,0.902} 
KAMBA     & 65.84         & 79.17        & 94.18              & 94.27        & 77.78            & 79.64 \\
\hline
\end{tabular}
\vspace{10pt} 
\caption{Ablation studies on the effect of introducing changes in the classical Mamba block on the model on BUSI dataset.}
\label{tab:mamba}
\end{table*}
We comprehensively evaluated our proposed KAN-Mamba FusionNet across various settings, to thoroughly examine and investigate the impact, effectiveness and potential of the proposed methods.\\

\noindent \textbf{Effect of KAMBA Block:} Table \ref{tab:mamba} shows the results of the ablation study to validate the effect of adding KAMBA block. We first evaluated classical Mamba with U-Net architecture. Then we replaced convolutional layers from classical Mamba with KAN. For the next experiment, we added BoA with the classical Mamba and KAN block. With each module addition, the table shows improved performance. Finally, our proposed KAMBA, which consists of SSM, KAN, BoA, spatial attention, and a skip connection, outperformed these models on the listed performance parameters. \\ 

\noindent \textbf{Effect of Bag of Activation Functions:} Table \ref{tab:ablation1} shows the effect of adding BoA in the KAMBA block in our proposed KAN-Mamba FusionNet model, compared to using no activation function or a single activation function. Introducing a single activation function (ReLU) results in improved performance of the model compared to using no activation function in the KAMBA block. Further, by incorporating a composition of multiple pre-defined activation functions, essentially a bag of activation (BoA) functions, the model's learning capability is further enhanced, leading to even better results. These results underscore the importance of adding BoA, demonstrating that the diversity of activation functions plays an important role in enhancing the performance of the model.

\begin{table*}[t]
\centering
\fontsize{8}{12}\selectfont
\begin{tabular}{|c|c|c|c|c|c|c|}
\hline
\multirow{2}{*}{\parbox{2cm}{\centering Activation Functions}} & \multicolumn{6}{c|}{Performance Metrics} \\
\cline{2-7}
                      & IoU$\uparrow$ & F1$\uparrow$ & Accuracy$\uparrow$ & AUC$\uparrow$ & Precision$\uparrow$ & Recall$\uparrow$ \\
\hline
None               & 62.73          & 77.43         & 93.43               & 92.04        & 74.14               & 75.22 \\
Single (ReLU)           & 64.61          & 78.26         & 93.91               & 92.12       & 77.59              & 74.87 \\

\rowcolor[rgb]{0.678,0.847,0.902}  
Proposed BoA    & 65.84         & 79.17        & 94.18              & 94.27        & 77.78           & 79.64 \\
\hline
\end{tabular}
\vspace{10pt} 
\caption{Ablation studies on the effect of bag of activation functions in KAMBA block of the model on BUSI dataset.}
\label{tab:ablation1}
\end{table*}
\section{Conclusion}
In this paper, we propose a novel KAN-Mamba FusionNet architecture to improve the medical image segmentation accuracy. As part of this architecture, we have designed KAMBA block with KAN, BoA, SSM, spatial attention module and a skip connection to better capture non-linearities and handle long-range dependencies in the input data. Our experiments across the BUSI, Kvasir-Seg, and GlaS datasets validate our hypothesis, demonstrating better IoU and F1 scores in comparison with the state-of-the-art methods. Furthermore, our ablation studies provide crucial insights, revealing the key role of each architectural component in the KAN-Mamba FusionNet model. In summary, our proposed approach considerably advances the overall accuracy, effectiveness and robustness of medical image segmentation with efficient computation. 

\bibliography{egbib}
\end{document}